# Measuring Similarity in Causal Graphs: A Framework for Semantic and Structural Analysis

Ning-Yuan Georgia Liu[1], Flower Yang[1], Mohammad S. Jalali[1, 2,*]

[1]MGH Institute for Technology Assessment, Harvard Medical School, Boston, Massachusetts, United States of America

[2]Sloan School of Management, Massachusetts Institute of Technology, Cambridge, Massachusetts, United States of America

***Corresponding author:**

Email: msjalali@mgh.harvard.edu

# Abstract

Causal graphs are commonly used to understand and model complex systems. Researchers often construct these graphs from different perspectives, leading to significant variations for the same problem. Comparing causal graphs is, therefore, essential for evaluating assumptions, integrating insights, and resolving disagreements. The rise of AI tools has further amplified this need, as they are increasingly used to generate hypothesized causal graphs by synthesizing information from various sources such as prior research and community inputs, providing the potential for automating and scaling causal modeling for complex systems. Similar to humans, these tools also produce inconsistent results across platforms, versions, and iterations. Despite its importance, research on causal graph comparison remains scarce. Existing methods often focus solely on structural similarities, assuming identical variable names, and fail to capture nuanced semantic relationships, which is essential for causal graph comparison. We address these gaps by investigating methods for comparing causal graphs from both semantic and structural perspectives. First, we reviewed over 40 existing metrics and, based on predefined criteria, selected nine for evaluation from two threads of machine learning: four semantic similarity metrics and five learning graph kernels. We discuss the usability of these metrics in simple examples to illustrate their strengths and limitations. We then generated a synthetic dataset of 2,000 causal graphs using generative AI based on a reference diagram. The synthetic dataset was designed to include a wide range of variations, from perfect matches to entirely unrelated diagrams, enabling us to evaluate how well the metrics captured differences, including the semantic relationships of variable names and graph structural similarities. Our findings reveal that each metric captures a different aspect of similarity, highlighting the need to use multiple metrics. This research contributes to advancing methods for causal graph comparison, with potential applications across diverse fields.

## Author Summary

Causal graphs are diagrams that help us understand how different factors are connected and influence one another, making them an essential tool for studying complex problems like climate change, public health, and social dynamics. However, comparing these graphs can be tricky. Differences in variable names, graph structures, or the way relationships are represented can make it hard to assess whether two graphs convey the same information or something completely different. In this study, we explore a variety of metrics to compare causal graphs, focusing on their structure and the meaning of their variable names. We tested these methods using synthetic datasets created by artificial intelligence, simulating what might happen if hundreds of people independently tried to map out the same complex system. Our results present a comparison of these metrics, highlighting their differences in evaluating semantic meaning and graph structure. As both human- and AI-generated causal graphs become more common, the need for a rigorous and effective evaluation grows. This study lays the groundwork for developing better tools to compare and interpret complex systems.

# Introduction

Understanding complex systems is a central challenge in scientific research. Causal graphs are a powerful tool for translating intricate cause-effect relationships in complex systems into simple graph representations [1]. They help researchers map out their hypotheses, refine mental models, and make sense of the complex connections between different factors.

Causal graphs are widely applied across disciplines with various methodologies. For instance, directed acyclic graphs (DAGs) are used in computer science and mathematics to model hierarchical structures and study dependency relationships, aiding problems in causal inference and probabilistic reasoning [2]. Causal loop diagrams (CLDs) are used in systems science for mapping feedback loops and understanding how they drive emergent behaviors in complex systems [3]. Similarly, structural equation modeling (SEM) is commonly used in social sciences to represent causal relationships between latent and observed variables [4]. Despite differences, they share a common objective: they aim to simplify complexity and depict causal relationships by representing variables as nodes and their causal interdependencies as directed edges through a graph structure.

Causal graphs reflect different perspectives and methodologies, leading to variations even when representing the same phenomenon, such as differing interpretations of environmental changes [5–7]. Therefore, comparing causal graphs built for the same phenomena is essential for evaluating assumptions, integrating insights, and resolving disagreements between different theoretical frameworks and mental models. Additionally, recent advancements in generative AI have further amplified this need. These tools can synthesize data inputs into causal structures derived from literature, participatory processes, or large datasets and have significant potential for automating and scaling causal modeling of complex phenomena [8,9]. However, the generation of causal graphs through AI models is inherently subject to variations driven by the underlying algorithms, data quality, model training, and contextual nuances. For example, using the same input data, AI-generated graphs may differ in how they define key variables, represent causal links, or emphasize certain relationships over others. Despite the growing reliance on these AI-generated graphs, no standardized framework exists to systematically compare and evaluate their created causal graphs. This gap hinders the ability to assess their reliability and validity.

A growing body of research in diverse domains has focused on developing metrics to measure the similarity between causal graphs. For example, metrics developed for DAGs have been proposed to compare causal graphs based on distances between causal effects, such as structural intervention distance [10], Kendall-tau distance [11], and separation distance [12], assuming node correspondence. Comparison metrics for CLDs focus on properties such as link polarity and feedback loops 13]. More recent work has even adapted from information retrieval metrics, such as precision, recall, and F1 scores, to measure the alignment of causal structures in CLDs [15]. Notably, we found no metrics for comparing SEM-based causal graphs despite their widespread use.

A key limitation across existing comparison metrics is the focus on structural aspects of causal graphs, assuming identical variable names and one-to-one node correspondence. This approach neglects common cases where semantically similar variables are described using different terms and where the number of nodes may differ. As a result, no existing research looks at both semantic and structural similarity when comparing causal graphs.

We address this gap by exploring existing comparison metrics, including semantic similarity [16] and graph comparison metrics [17,18]. Based on our predefined criteria for assessing causal graphs, we narrowed our focus and selected nine similarity metrics. Finally, we assess the usefulness of the selected metrics by using a set of synthetic datasets generated by a large language model (LLM).

# Material and Methods

We draw on similarity metrics from two areas of machine learning: semantic similarity measures to assess how closely variable names are related in meaning, and graph comparison metrics to evaluate the structural similarities of causal graphs. Here, we discuss these metrics and the study design.

## Semantic similarity between variable names

Semantic similarity, also known as semantic measures, are techniques within the broader field of natural language processing (NLP) developed to estimate how closely linguistic entities, such as words, sentences, or concepts, align in their underlying semantic meaning [16,19]. They capture the degree of semantic relatedness between entities by representing them with numerical scores.

We explore how semantic similarity measures can aid in comparing causal graphs, where variable names might differ in wording but share the same underlying concept.

### Selection criteria for semantic similarity measures

A diversity of approaches for measuring semantic similarity have been proposed. We reviewed and examined over 40 different semantic measures from [16] based on factors such as the type of linguistic units the measures aim to compare, the source of information used to characterize the compared elements, and the underlying assumptions of the mathematical approach, such as calculating word similarity by geometric approach or by the number of overlapping sets of words.

Given the focus of this study is on the semantic similarity of the variable names in causal graphs, we considered four selection criteria: (i) Independence from external ontologies: Our analysis does not rely on pre-existing structured knowledge bases, i.e., the selected measures should not require domain-specific ontologies. (ii) Suitability for short textual inputs: As variable names in causal graphs are typically concise, e.g., consisting of a few words rather than a full sentence, the chosen measures must handle such brief segments. (iii) Interpretability of similarity scores: The measures should provide intuitive insights into the semantic relationships among variable names, ensuring that the results can be utilized in subsequent analyses. (iv) Computational efficiency: The measures should be computable in a reasonable time without major resource needs to ensure usability and scalability.

### Selected semantic similarity measures

We applied the predefined selection criterion (i-iv) to the initial set of 40 measures. Out of the 40 measures, 20 rely on external ontologies (i.e., knowledge-based measures), conflicting with criterion (i), and were therefore excluded. The remaining measures, which are corpus-based, i.e., deriving semantic relationships from statistical patterns in textual data, were deemed appropriate for brief and domain-agnostic variable names in causal graphs. Among the 20 corpus-based measures, we focused on distributional-based approaches, as they are particularly relevant for comparing short textual inputs (satisfying our criterion (ii)). Therefore, we focused on the distributional-based measures, grouping them into three conceptual approaches: geometric or spatial, set-based, and probabilistic. Each metric in this collection comes in different variations when measuring semantic relatedness, including capturing the multi-dimensional context of words, assessing their relative positions in semantic space, and computing the probability of their cooccurrence within a corpus [20–24]. From the 20 measures reviewed, we selected four metrics that met our selection criteria and used them as representation examples. These measures are summarized in Table 1.

**Table 1.** Characteristics of the selected semantic similarity metrics

| Metrics | Approach | Characteristics and considerations | Scale* | References |
|---|---|---|---|---|
| M1: BLEU | Probabilistic | Identifies surface-level structural similarities in phrases by measuring the overlap of word sequences. Suitable for assessing syntactic fidelity; however, it does not account for the meaning or order of words. | 0 to 1 | [25] |
| M2: Fuzzy Matching | Set-based | Also known as approximate string matching, determines the closeness of two strings using the Levenshtein edit distance. Suitable for capturing partial matches, typographical errors, and minor discrepancies in variable names; prone to produce higher false positives. | 0 to 1 | [26] |
| M3: Cosine Similarity | Geometric | Measures similarity by calculating the cosine of the angle between word vectors and determines whether the two vectors point in the same direction. Suitable for identifying rephrased or synonymously expressed variables; however, it is sensitive to text length and performs poorly in sparse data. | 0 to 1 | [27] |
| M4: Negative Euclidean Distance | Geometric | Measures similarity by computing the straight-line distance between word vector embeddings. Suitable for analyzing spatial relationships in embedding space; however, in high-dimensional spaces, it may lose meaningfulness due to data sparsity. | $-\infty$ to 0 | [28] |

*Higher values indicate greater semantic similarity; note that M4 has a different scale.

To compare the performance of these four measures, we first embed the variable phrases using the Sentence-BERT (*all-mpnet-base-v2*) model [29]. Sentence-BERT is a widely used pre-trained general-purpose language model that generates semantic embeddings, making it ideal for measuring the semantic similarity of short text inputs. This model is also commonly used as a baseline for tasks like semantic similarity within the NLP community. Other text embedding models, such as word2vec [30], GloVe [31], ELMo [32], and Universal Sentence Encoder [33], could also be employed, each leveraging different techniques to capture linguistic meanings.

## Graph comparison between causal graphs structure

Graph comparison is a subfield of graph theory and network analysis that focuses on identifying similarities between two or more graph structures [18]. A graph consists of a set of vertices and edges, where both vertices and edges can carry various attributes, such as labels, weights, or directions. As a versatile data structure, graphs are widely used to analyze and encode structural information. Applications include social network analysis (examining relationships among individuals and groups), molecular cell analysis (understanding cell interactions), and chemical composition analysis (identifying structural similarities among compounds).

We extend the use of graph comparison approaches to causal graphs, a specialized type of graph-structured data where the nodes (also known as "vertices" in graph theory) represent variables, and edges indicate causal relationships, including their direction and polarity (positive or negative). For example, Figure 1 illustrates a causal graph, a CLD, where variables such as *student enrollment, school capacity strain,* and *school ranking* appear as nodes. Meanwhile, the edges (labeled with + or -) capture the nature of their causal directions—positive represents a change in the same direction, while negative represents the opposite.

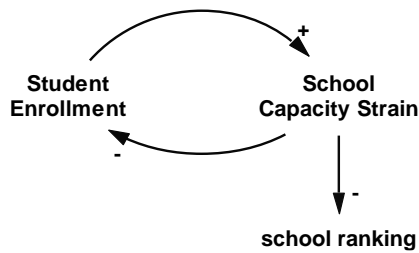

**Figure** 1**.** Example of a causal graph, specifically a causal loop diagram

## Selection criteria for graph comparison measures

Numerous graph comparison metrics have been proposed to address the diverse types, sizes, attributes, and structures (e.g., flow-based or cyclic) of graphs. They are categorized into three groups based on their underlying approaches for calculating distances between graphs to assess similarity [18]: spectral distance measures [34], matrix distance measures [35], and learning graph kernels [17].

Focusing on structural comparison for causal graphs, we established three selection criteria to identify metrics suitable for comparing causal graphs. These criteria are: (i) Computational feasibility and efficiency: Given the high computational costs associated with graph comparison and the feasibility of existing software, the metric must be capable of efficiently analyzing multiple causal graphs to ensure practical applicability and ease of implementation. (ii) Node correspondence independence: Causal graphs frequently vary in size and structure, which makes strict one-to-one node mapping infeasible. The selected measures must function independently of node correspondence to accommodate the structural diversity of causal graphs. (iii) Attribute sensitivity: The ability to incorporate node names and edge labels is crucial, as these attributes encode meaningful causal relationships necessary for understanding the structure of causal graphs.

## Selected graph comparison measures

Considering these criteria, we excluded two categories of graph comparison measures: spectral distance and matrix distance measures. They are computationally expensive, failing to meet criterion (i). Also, matrix distance measures require strict node correspondence, failing to meet criterion (ii).

In contrast, learning graph kernels have emerged as a suitable approach for comparing causal graphs, as they fulfill all our three criteria. Recently introduced in machine learning [17], these kernels work by embedding a set of graphs into the Euclidean space and then computing a measure of similarity between the embedded graphs. They come in various forms, measuring similarity by analyzing structural features such as subgraphs, shortest paths, and node or edge distributions. Furthermore, their independence from node correspondence, along with their support for labeled nodes, edges attributes, and directed edges, makes them ideal for causal graph comparison. Learning graph kernels also have multiple extensions based on their focus on the structural aspects of graphs, their ability to handle node or edge attributes, and whether they use explicit or implicit computation schemes [36–38].

From the 20 learning graph kernels reviewed by Nikolentzos et al. [17], we selected five with the strongest alignment with our criteria. The selected methods and their key characteristics are summarized in Table 2.

**Table 2**. Characteristics of the selected learning graph kernel metrics

| Metrics | Characteristics and considerations | Scale* | References |
|---|---|---|---|
| G1: Pyramid match | Employs a multi-resolution histogram pyramid in the feature space to form an approximate partial matching between two sets of feature vectors. This kernel does not assume independence and therefore no explicit match search is required.<br><br>It is robust to clutter since it does not penalize the presence of extra features; however, by aggregating features at multi-resolutions, this abstraction may hinder the kernel's ability to distinguish between graphs that are structurally similar at higher levels but differ in specific local nuances. | 0 to 1 | [39,40] |
| G2: Shortest path | Decomposes graphs into shortest paths and counts the distance between two graphs based on path lengths and endpoint node labels.<br><br>Ideal for capturing significant causal paths and relationships, particularly for cyclical graph structure, which avoids tottering;** however, the presence of multiple shortest paths between two nodes increases computational complexity as the graph size grows. | 0 to 1 | [41] |
| G3: Subgraph matching | Counts the number of matchings between subgraphs of bounded size in two graphs while considering the node labels.<br><br>Effective for precise structural comparisons; however, it can be computationally expensive when the number of subgraphs grows rapidly in graphs with diverse labels. | 0 to 1 | [42] |
| G4: Weisfeiler-Lehman (WL) vertex histogram | Measures graph similarity by iteratively capturing how its neighbors influence each labeled node, thereby encoding richer topological and node label information.<br><br>It is fast and scalable for analyzing large, node-labeled graphs and focuses on node labels and structural information; however, it does not account for edge labels. | 0 to 1 | [43,44] |
| G5: WL edge histogram | Iteratively captures the influence of neighboring edge labels, thereby encoding richer structural and edge label information. | 0 to 1 | [43,44] |

| Metrics | Characteristics and considerations | Scale* | References |
|---------|-----------------------------------|--------|------------|
|         | It is fast and scalable for processing large, edge-labeled graphs; however, it focuses solely on edge labels and structural information without accounting for node labels. |        |            |

*Higher values indicate greater structural similarity.
**Tottering is a phenomenon where a walk (a sequence of nodes traversed through a graph) repeatedly visits the same nodes in a cycle, leading to artificially high similarity scores between graphs, even if they are not structurally similar [45].

# Examples and synthetic data for metric analysis

For analysis, we first demonstrate how each selected metric functions using a set of simple causal graphs. Next, we generate a synthetic dataset of causal graphs using LLMs to examine how the selected metrics perform across a diverse range of graph variations. We use CLDs as our exemplar, as their cyclic structures and labeled edges make them particularly well-suited for this evaluation and an ideal testbed for assessing our metrics.

## Simple examples of metric sensitivity

In the Results section, we will compare a simple reference graph (Figure 1) against multiple comparative graphs to illustrate the behavior of the selected metrics. Specifically, we demonstrate the semantic similarity metrics by gradually modifying the variable names, while the graph comparison metrics are showcased through incremental changes—such as adding, deleting, or altering the polarity of edges.

## Synthetic data for exploring metric sensitivity

To evaluate the performance of the selected metrics, we generated a synthetic dataset of CLDs using LLMs, which offer a practical solution to the scarcity of real-world data through synthetic data generation [46]. The dataset was created using a simplified representation of the Limits to Growth model (Figure 2). First introduced in 1972 [47], the Limits to Growth model illustrates how population expansion becomes unsustainable as resources diminish, ultimately leading to a decline. It has been widely studied across disciplines, including environmental science, sociology, and other fields [48,49], offering insights into the causal relationships between population growth and resource constraints.

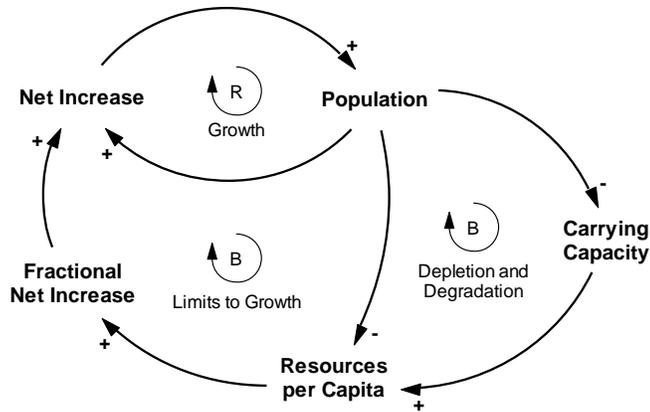

**Figure** 2. Limits to Growth causal loop diagram; adapted from [3]
Note: The labels inside the cycles represent the feedback loops, where R indicates a reinforcing loop while B indicates a balancing loop.

We began by drafting a descriptive paragraph that outlined the core causal relationships of the Limits to Growth model, adapted from [3] (see Supplementary Document). This paragraph was provided as input to the LLM (*GPT-4o-mini* model [50], performed in December 2024) to generate a synthetic dataset of 2,000 CLDs. To ensure the synthetic data quality [46], we employed an iterative refinement process that combined few-shot prompting [51] (providing examples and detailed instructions to guide the LLM in constructing diverse CLDs) and curated prompts [8] (designed to extract the elements of a causal graph, e.g., variable names and causal relationships).

We aimed to generate a wide range of quality variations, ranging from causal graphs that were identical to the reference CLD (Figure 2) to those that were somewhat similar, poorly aligned, or entirely unrelated. This approach simulates the task of asking 2,000 individuals to independently create CLDs based on a textual description. Importantly, the diversity of the generated data was intentional in testing whether the selected metrics could capture a full spectrum of similarities and differences. Unlike other uses of synthetic data, these 2,000 generated causal graphs were not meant to be uniformly high quality—instead, the goal was to create variability, not accuracy.

Detailed information about the dataset generation process and its evaluation can be found in the Supplementary Document. The metrics calculations were performed in Python, and the code is available in the project GitHub repository—see Data Availability statement.

# Results

This section presents the results of our analysis, beginning with a demonstration of how the selected metrics respond to controlled modifications in simple examples, followed by an evaluation of their behavior on the synthetic dataset.

## Metric sensitivity in simple examples

### Semantic similarity metrics

In Table 3, we present five simple examples compared to a reference graph, evaluated using the semantic similarity metrics detailed in Table 1: BLEU (M1), Fuzzy Matching (M2), Cosine Similarity (M3), and Negative Euclidean Distance (M4).

Each row demonstrates how the semantic similarity scores vary as the variable name *"student enrollment"* is altered (e.g., *"student demand," "student attendance"*), with these changes highlighted in red. When the two graphs are identical (as shown in the first row), M1-M3 yield a perfect score of 1, while M4 (being on a different scale) returns a perfect score of 0. However, in subsequent rows, the metric values exhibit a mixed pattern, which we discuss in the following section.

**Table** 3. Semantic similarity scores for variations of a reference causal loop diagram

| Reference graph | Comparative graph | Semantic similarity score* | | | |
|---|---|---|---|---|---|
| | | M1 | M2 | M3 | M4 |
| Student Enrollment → School Capacity Strain (+/−) | Student Enrollment → School Capacity Strain (+/−) | 1 | 1 | 1 | 0 |
| | Student Demand → School Capacity Strain (+/−) | 0.29 | 0.87 | 0.95 | -0.33 |
| | Student Attendance → School Capacity Strain (+/−) | 0.29 | 0.87 | 0.95 | -0.33 |
| | Number of Students → School Capacity Strain (+/−) | 0.29 | 0.73 | 0.95 | -0.27 |
| | Number of Ice Creams → School Capacity Strain (+/−) | 0.16 | 0.64 | 0.73 | -0.73 |

\* M1 is BLEU (representing word sequence overlap); M2 is Fuzzy Matching (representing typological differences); M3 is Cosine Similarity (representing the cosine angle between word vectors); and M4 is Negative Euclidean Distance (representing the distance between word embeddings).
Note: The darker the color, the higher the similarity score.

## Graph comparison metrics

Table 4 presents six examples illustrating how graph similarity scores vary when intentional differences are introduced compared to a reference graph. The graph similarity scores are calculated using the selected graph comparison metrics detailed in Table 2. Specifically, we employ the following learning graph kernels: Pyramid match (G1), Shortest path (G2), Subgraph matching (G3), WL vertex histogram (G4), and WL edge histogram (G5).

In the first example (as shown in the first row), the comparison graph is identical to the reference graph, resulting in a perfect similarity score of 1 across all metrics, a condition known as graph isomorphism [52]. In subsequent examples, the similarity scores decrease in response to modifications in node and edge attributes (highlighted in red), demonstrating each metric's sensitivity to specific types of structural differences.

**Table 4.** Graph similarity scores for variations of a reference causal loop diagram

| Reference graph | Comparative graph | Graph similarity score* | | | | |
|---|---|---|---|---|---|---|
| | | G1 | G2 | G3 | G4 | G5 |
| Student Enrollment → School Capacity Strain (+), School Capacity Strain → Student Enrollment (-), School Capacity Strain → school ranking (-) | Student Enrollment ↔ School Capacity Strain; School Capacity Strain → school ranking | 1 | 1 | 1 | 1 | 1 |
| | (with red + edge variation) | 1 | 0.5 | 1 | 0.39 | 0.8 |
| | Demographic Student Population → Student Enrollment (+); Student Enrollment ↔ School Capacity Strain; School Capacity Strain → school ranking | 0.46 | 0.38 | 0.76 | 0.59 | 0.71 |
| | Demographic Student Population → Student Enrollment (+); Student Enrollment → School Capacity Strain (-); School Capacity Strain → Student Enrollment (-); School Capacity Strain → school ranking (-) | 0.46 | 0.19 | 0.76 | 0.32 | 0.45 |
| | Student Enrollment ↔ School Capacity Strain (without school ranking) | 0.33 | 0.29 | 0.58 | 0.35 | 0.95 |
| | Student Enrollment → School Capacity Strain (-); School Capacity Strain → Student Enrollment (-) | 0.54 | 0.35 | 0.87 | 0.08 | 0.45 |

\* G1 is Pyramid match (representing partial matching by employing multi-resolution histogram pyramid in the feature space); G2 is shortest path (representing the shortest path distance between graphs); G3 is Subgraph matching (representing the number of subgraph matching); G4 is WL vertex histogram (considered the neighbor node label and topological information); and G5 is WL edge histogram (considered the neighbor edge label and topological information).

Note: The darker the color, the higher the similarity score.

# Metrics sensitivity in synthetic data

## Semantic similarity metrics

To evaluate the usefulness of the metrics, Figure 3 presents the distribution of similarity scores for the four selected semantic similarity metrics: BLEU (M1), Fuzzy Matching (M2), Cosine Similarity (M3), and Negative Euclidean Distance (M4). Given our intentional generation of a wide range of causal graph variations, we expect a broad spread of similarity scores across the dataset. The distribution plot highlights how each metric captures the semantic similarities of variable names between the reference CLD and the 2,000 synthetic CLDs.

M1 demonstrates a skewed distribution toward lower similarity scores, indicating limited text overlap between variable names. This limitation arises from M1's reliance on exact word sequences, which restricts its ability to capture lexical variations. Metric M2 reflects a distribution with a greater spread and a wider range of similarity scores. This pattern shows M2's ability to identify partial overlaps and accommodate minor variations or typographical errors.

Metric M3 shows a strong concentration of higher scores, capturing conceptual relationships between variable names. By leveraging word embeddings, M3 identifies semantic connections even when the lexical forms differ significantly. Finally, M4, which measures straight-line distances in embedding space, produces a broad distribution of similarity scores (between -1.1 and 0 in this dataset). While capturing a wide range of variation, its reliance on magnitude and lack of a fixed range make it less interpretable than the other metrics.

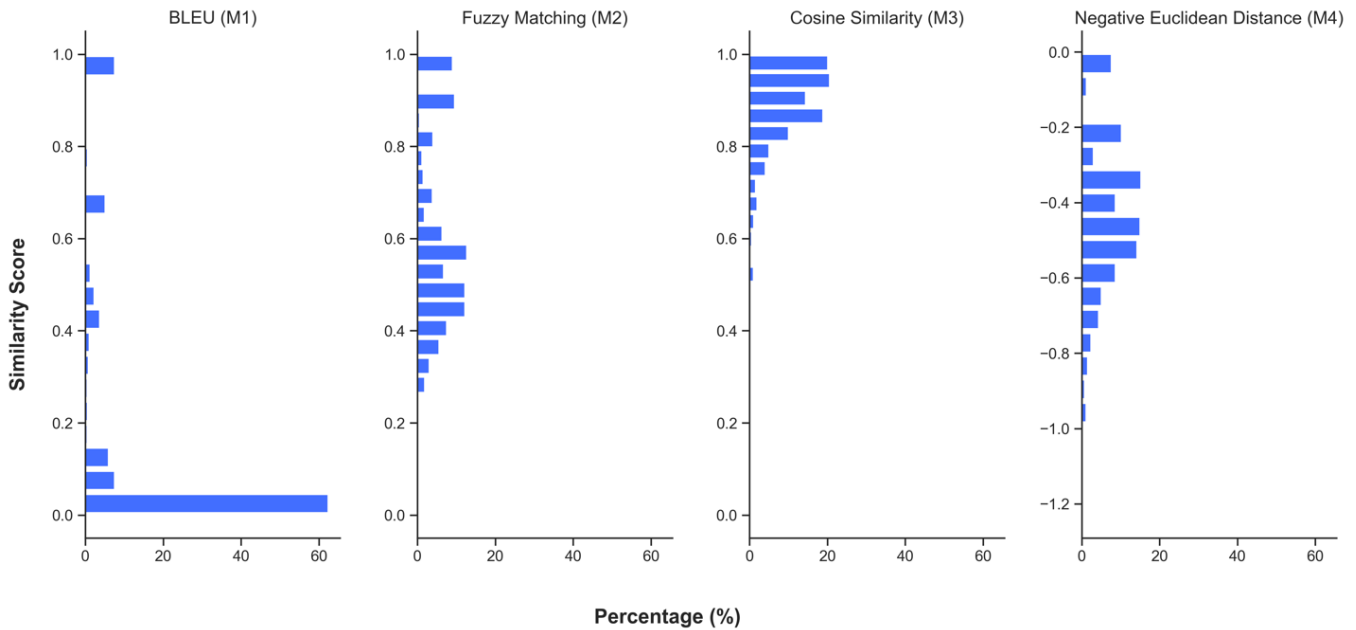

**Figure 3.** Semantic similarity metrics distribution on the 2,000 generated CLDs
Note: the range of Negative Euclidean Distance is different from the other three measures.

## Graph comparison metrics

Figure 4 presents the distribution of similarity scores for the five selected graph comparison metrics, which are specifically learning graph kernels, including the Pyramid Match (G1), Shortest Path (G2), Subgraph Matching (G3), WL Vertex Histogram (G4), and the WL Edge Histogram (G5). These scores reveal how each metric captures the structural similarity between the synthetic CLD and the reference CLD.

In Figure 4 G1 exhibits a skewed distribution toward the lower percentile because it calculates the similarity by measuring the partial matches between the two sets. G2 produces distributions heavily concentrated near 0, with most scores falling in the lower range. Metric G3, in contrast, presents a more dispersed distribution of similarity scores. By focusing on identifying exact subgraph matches, this metric captures partial structural overlaps, even when the overall graph alignment is imperfect. This flexibility makes G3 less restrictive than other metrics, as it accounts for shared substructures despite variations in node or edge labels.

Metric G4 evaluates the similarity between graphs by only considering node labels, which produces similarity scores of 0 for most comparisons due to mismatched variable names. However, in the few cases where node labels match exactly, G4 achieves a similarity score of 1. Conversely, the G5 focuses exclusively on edge labels while disregarding node labels, resulting in higher similarity scores for graphs with matching edge labels. While this emphasis on edge-level similarities helps identify closely related causal link patterns, it can be misleading as it overlooks broader structural differences and variations in node labels.

These observed patterns reflect the diverse characteristics of the graph comparison metrics, particularly their ability to handle structural differences, label distributions, and neighboring relationships.

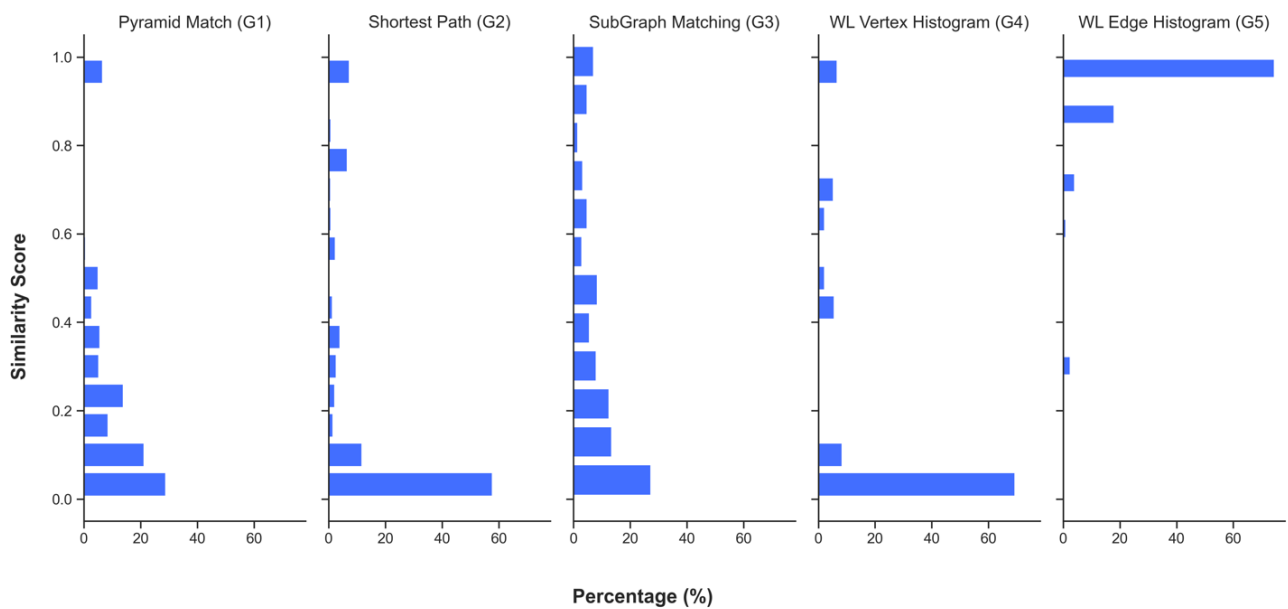

**Figure** 4. Graph comparison metrics distribution on the 2,000 generated CLDs

# Interpretation of the selected metrics

Here, we discuss how the metrics behave when comparing the reference CLD against the synthetic CLDs. Table 5 presents four CLD examples from the 2,000 synthetic CLDs, representing identical, strongly similar, moderately similar, and dissimilar graphs. While an ideal setup would involve controlled, stepwise modifications (e.g., a single variable change or an altered causal link), the nature of our generated dataset does not allow for such precise adjustments. Instead, these examples reflect the natural variation present in our set of generated CLDs, capturing a mix of differences in variable names, missing or extra nodes, and structural changes.

In the first case, all similarity scores are perfect since the two graphs are identical. In the second case, semantic similarity scores remain high because the variable names are identical between the two graphs, with no missing or altered labels. However, the graph similarity scores are lower, albeit relatively high, due to structural differences, including a missing causal link (*Population → Carrying Capacity*) and a reversed causal link (*Resources per Capital → Carrying Capacity*).

In the third case, semantic similarity scores get lower due to the absence of two variables, *Carrying Capacity* and *Net Increase*. Despite this, M2 and M3 remain high at 0.78 and 0.96, reflecting their ability to capture partial matches and synonymous meanings among the remaining variables. This is expected; M2 tolerates small textual variations, while M3 captures conceptual relationships, even when some variables are missing. In contrast, graph similarity scores (G1-G4) dropped as the structure diverged significantly. Metrics like G1 and G3 penalize missing or rearranged causal relationships, leading to lower scores. The absence of growth and depletion cycles markedly impacts structural alignment, with G3 yielding a score of 0.5 since it computes similarity by identifying matching subgraphs within the synthetic CLD. Meanwhile, G5 remains high (0.99) because it focuses solely on edge labels. This makes sense as G5 evaluates only edge labels and ignores node labels entirely. However, such result can be misleading, as the metric does not account for label node differences between the two graphs.

In the final case, both semantic and graph similarity scores are low, as the synthetic CLD differs significantly from the reference CLD in terms of variable names and causal structure. Specifically, the variable name *Net Increase* is changed to *Increase*, and a new variable name, *Growth Factor* is introduced. M1 yields a score of 0 as no overlapped word sequence is found. M2 yields a score of 0.32, while M3 achieves 0.65, reflecting the partial retention of terms related to growth and increase. M4 yields a score of -0.83 due to the large Euclidean distance between the two-word vectors. The graph similarity scores are uniformly zero across G1-G4, as neither the node labels nor the graph structure are found. However, G5 yields a score of 0.95 since it ignores node labels and only considers edge labels, thereby identifying one cyclic structure with all positive signs, which again provides a misleading result.

These examples highlight the metrics' varying sensitivities to variable names, causal links, and graph structure changes. This analysis underscores the importance of selecting metrics based on the specific requirements of a given task, whether for identifying exact matches, capturing partial alignments, or evaluating nuanced semantic and structural relationships.

**Table 5.** Semantic similarity and graph similarity scores on synthetic CLD examples compare to the reference CLD

| Reference CLD | Synthetic CLD | Semantic similarity score* | | | | Graph similarity score* | | | | |
|---|---|---|---|---|---|---|---|---|---|---|
| | | M1 | M2 | M3 | M4 | G1 | G2 | G3 | G4 | G5 |
| 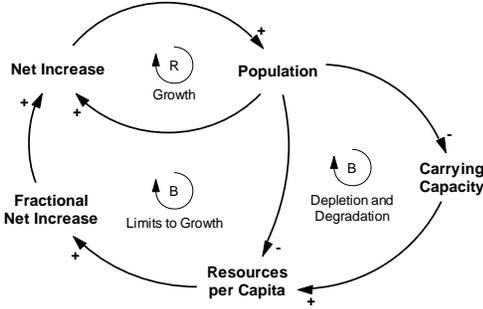 | 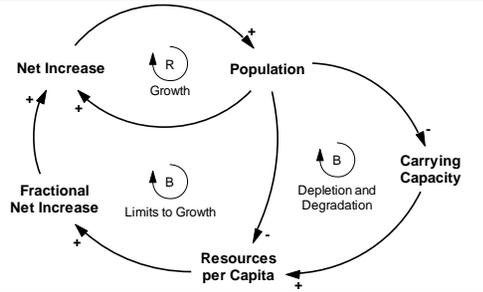 | 1 | 1 | 1 | 0 | 1 | 1 | 1 | 1 | 1 |
| | 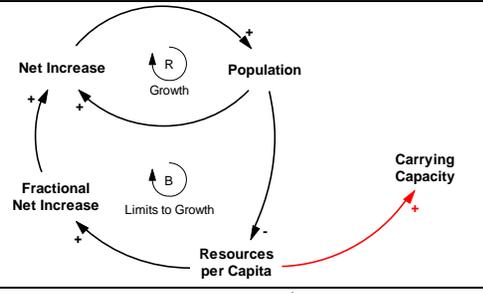 | 1 | 1 | 1 | 0 | 0.48 | 0.73 | 0.94 | 0.64 | 0.99 |
| | 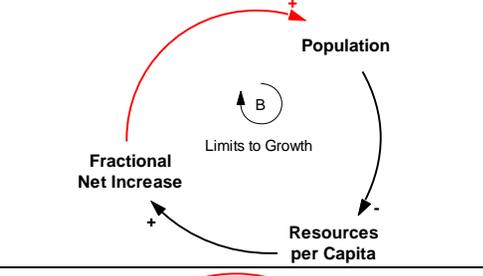 | 0.36 | 0.78 | 0.96 | -0.28 | 0.22 | 0.27 | 0.5 | 0.05 | 0.99 |
| | 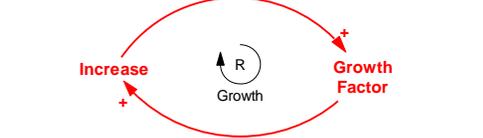 | 0 | 0.32 | 0.65 | -0.83 | 0 | 0 | 0 | 0 | 0.95 |

## Discussion

This study reveals how different metrics capture similarities between causal graphs, emphasizing their distinct strengths and limitations.

For semantic similarity, Cosine Similarity consistently provided high scores, reflecting its ability to capture conceptual relationships even when variable names diverged lexically. However, this can be a double-edged sword—while it effectively recognizes synonymous terms, it may also assign high similarity to conceptually related but distinct terms, which could be misleading in certain contexts. We observed this behavior in cases where two variable names shared a general theme (e.g., "Net Increase" vs. "Increase") but had different implications within the causal structure. In contrast, BLEU showed a significant limitation in detecting deeper semantic connections due to its dependence on exact word sequence matching. This suggests that while BLEU might be suitable for tasks requiring high syntactic fidelity, it underperforms when variable names are rephrased or semantically aligned but syntactically distinct. Fuzzy Matching, with its sensitivity to partial overlaps, emerged as a versatile tool for detecting subtle variations in wording, making it particularly useful for causal graphs where minor changes in terminology matter. However, its ability to recognize partial matches can also introduce noise when variable names differ meaningfully rather than superficially. Finally, Negative Euclidean Distance produced a broad range of similarity scores as well, but its reliance on magnitude made direct comparisons across different datasets less intuitive.

On the structural side, graph comparison metrics exhibited notable variations in sensitivity to different graph features. Pyramid Match, Shortest Path, and Subgraph Matching emerged as useful metrics for identifying structural similarities in graphs with shared subcomponents, even when overall topological alignment was imperfect. Subgraph Matching, in particular, demonstrated an ability to capture partial matches, which is critical when comparing graphs that retain shared causal mechanisms but differ in complexity or granularity. Additionally, metrics such as Shortest Path were effective in graphs with cyclical structures, as they count the shortest linear paths between two endpoints of nodes, which is ideal when feedback loops are present. The two WL metrics revealed an intriguing trade-off: WL Vertex Histogram captured node-based similarities, but its disregard for edge labels limited its applicability for graphs where edge polarity or directionality is critical. Similarly, the WL Edge Histogram prioritized edge-level features while ignoring node labels, highlighting the challenge of balancing structural and attribute-level comparisons.

Interestingly, the divergence among both semantic and graph comparison metric performance underscores the inherent complexity of comparing causal graphs. Graphs that closely align semantically but differ structurally, or vice versa, present unique challenges that no single metric fully resolves. This is particularly evident in cases where causal links are preserved but labeled differently, leading to high semantic similarity yet low structural similarity scores. Conversely, certain structural metrics failed to recognize meaningful alignment when variable names differed, demonstrating their reliance on strict node naming alignment. These discrepancies reveal the need for careful consideration when selecting metrics for specific tasks, as each metric emphasizes different facets of graph similarity. For example, some applications may require prioritizing structural alignment over semantic similarity, while others—such as text-mined causal graphs [53]—may demand a stronger emphasis on semantic consistency. This nuanced understanding is crucial for applications such as evaluating AI-generated causal graphs, where both semantic fidelity and structural integrity are critical to validating their reliability.

One avenue for future research is the integration of semantic and structural metrics into a unified framework. Graph similarity metrics employed in our analysis rely on the strict alignment of node names without considering the semantic relationships between different node labels. This limitation raises an important question: Can the two types of metrics be integrated into a single measure that accounts for both semantic and structural characteristics? Moreover, given the integrated metric, the diverse components that constitute a causal graph, such as the semantics of variable names, the node and edge labels, and the directionality of edges, introduce another complex task of determining the relative weighting of each component's importance for similarity. The integration of these metrics likely depends on the specific needs of the project. For example, in a case like the Limits to Growth model, where key causal loops define system behavior, a combined metric could prioritize structural alignment by assigning higher weights to correctly matched feedback loops while still considering semantic similarity to accommodate variations in variable names. Developing such a framework would require careful calibration to ensure that both dimensions contribute meaningfully to the comparison. Additionally, empirical studies could further explore how domain-specific applications, such as epidemiology or economics, weigh these factors differently and whether customized metric weightings improve interpretability and utility.

A key limitation of this study is the constrained scope of graph type and size. The Limits to Growth model represents a relatively small causal graph with five nodes and six causal links. Extending this research to include larger graphs would allow testing metrics' scalability and computational efficiency, ensuring reliability in more complex and varied applications. Additionally, our study exclusively examines CLDs, which primarily exhibit cyclic patterns. Future research should also consider other types of causal graphs that feature different structural characteristics and variable naming conventions.

Another limitation is that the synthetic dataset was generated using LLMs, introducing variability that was not precisely controlled. While this aligns with real-world AI-generated causal graphs, it makes it difficult to isolate the exact impact of specific changes on similarity scores. Future work could explore more systematically controlled datasets to provide finer-grained insights into how each metric responds to distinct modifications. Additionally, the extent to which LLM-generated graphs align with expert-designed causal models remains an open question, requiring further validation of their interpretability and reliability.

Finally, the computational feasibility of these metrics for large-scale causal graphs remains an open question. Some methods, particularly those relying on graph kernels, may not scale well to larger, more complex systems. Future studies should assess whether these approaches can be optimized or if alternative techniques, such as deep learning-based embeddings, offer better scalability for high-dimensional causal modeling.

Despite these limitations, this study represents a first step toward systematically evaluating and applying metrics for comparing causal graphs. By examining both semantic and structural approaches it provides a foundation for comparing different representations of complexity, helping advance the tools needed to analyze and interpret causal structures in complex systems.

Finally, as this research area continues to evolve, numerous variations and extensions of these metrics are being developed. While the selected metrics provide a foundational framework, they are by no means the most advanced. Future research should explore more sophisticated and specialized metrics for causal graph comparison to improve applicability and scalability.

**Acknowledgments**: We thank Ali Akhavan Anvari, Mariya Andreeva, Zeynep Hasgul, and participants at the Learning Club at MGH Institute for Technology Assessment for their constructive feedback on earlier versions of this analysis.

**Data Availability**: Data and analysis code are available in the Supporting Information and online repository available at https://github.com/georgia-max/Causalgraph-comparison.

**Funding**: None

**Competing interests**: The authors have declared that no competing interests exist.

# Supporting Information

# SI1. Synthetic data generation process

We employed an LLM to generate a set of synthetic causal loop diagrams (CLDs) of the Limits to Growth model by following a similar data generation approach documented in [1], which includes the descriptive paragraph as an input, the few-shots prompts, and the LLM model setup.

## Limits to growth model descriptive paragraph

The first step is to draft a descriptive paragraph that captures the causal relationships of the Limits to Growth model, adapted from [2]. It served as the basis for generating synthetic CLDs and is designed to encapsulate the feedback mechanisms governing population growth and resource availability. The descriptive paragraph that feeds into the LLMs is as follows:

> *Population grows based on the net increase, which is influenced by the fractional net increase—the rate at which the population grows per person. This rate isn't constant; it depends on the resources available per person. When resources are plentiful, the rate remains high, allowing growth to continue. However, as resources are stretched thinner with a growing population, the rate slows down, reducing the pace of growth. Additionally, the growing population depletes and degrades these resources, creating a cycle where fewer resources lead to slower growth, ultimately limiting how much the population can expand.*

## Few-shot prompts setup

We employed a few-shot prompting approach [3] to guide the LLM and improve model performance by demonstrating the desired input-output behavior through structured examples. This method provides the LLM context and examples illustrating the expected relationships between input text and the generated CLDs. The few-shot prompt consisted of three key components: i) A short instruction setting the task context, e.g., "*You are a helpful assistant that generates synthetic data of causal loop diagrams based on the given hypothesis.*"; ii) A curated set of 15 examples, each containing a descriptive paragraph, a CLD example, and a similarity score graded by a human modeler, indicating the quality of the CLD. And iii) Stepwise instructions specifying the required format and structure of the generated output.

This few-shot prompt setup ensured that the generated CLDs adhered to meaningful causal relationships while allowing diversity in semantic names and structure. Additionally, we explicitly guided the LLM in structuring its output by specifying the key elements of a causal graph in the final prompt. These included variable extraction, causal relationships, and structural variations to ensure a diverse set of graph structures, node labels, and causal links in the output dataset. The exact prompts can be found in the online data repository.

## Model specification

We employed the *GPT-4o-mini* LLM model [4] (December 2024) for the synthetic data generation process. We randomly sampled the model's *temperature* between 0.8 and 1.2 at each iteration, as suggested by [5], to enhance diversity in text data generation. The *temperature* parameter controls token sampling probabilities: a higher temperature increases the model's creativity by encouraging more diverse outputs, while a lower temperature produces more deterministic responses [6].

## SI2. Synthetic dataset evaluation

Evaluating the synthetic dataset generated by the LLM can be challenging [7]. Ideally, a human-based evaluation would be necessary; however, as the sample size increases, so does the labor cost of manual assessment. To address this, we followed [7] for evaluating synthetic datasets with a particular focus on data diversity. It is important to note that, unlike many synthetic dataset generation tasks that prioritize accuracy, our objective for this study is to produce a diverse set of CLDs from textual descriptions.

To assess the diversity of the structural properties in the synthetic dataset, we compare six common graph-level characteristics of the synthetic dataset against the reference CLD, as shown in S1 Table. The metrics include the number of variables (nodes), causal links (edges), cycles, graph density, transitivity, and connectivity. Graph density measures the ratio of actual causal links to the maximum possible links, providing insight into the overall connectivity of the CLD. A higher density suggests a more interconnected causal structure, whereas a lower density indicates sparsely connected variables. Graph density is as follows:

$$D = \frac{m}{n(n-1)}$$

where n is the number of nodes and m is the number of edges in the CLD.

Transitivity computes the fraction of all possible triangles present in a CLD, which is used to identify feedback loops, a key component in CLDs. Transitivity is as follows:

$$T = \frac{3t}{a}$$

where t is the number of triangles, and a is the number of triads (two edges with a shared node).

Connectivity [8] is the average of local node connectivity over all pairs of nodes of a graph. This metric assesses the extent to which the nodes within a CLD remain linked, ensuring that the generated causal structures maintain coherence and completeness. Higher connectivity suggests a well-integrated graph, whereas lower connectivity may indicate a fragmented or incomplete graph. The average connectivity is as follows:

$$\bar{k}(G) = \frac{\sum_{u,v} k_G(u,v)}{\binom{n}{2}}$$

where $k_G(u,v)$ is the local node connectivity of graph G (i.e., CLD), and n is the number of nodes.

The results in S1 Table show that the synthetic dataset effectively captures the overall distribution of these structural properties compared to the reference CLD. This suggests that the generated dataset maintains a diverse representation of CLDs, aligning with the intended goal of promoting diversity within the dataset.

**S1 Table.** Graph-level characteristics of the synthetic dataset

| Metrics | Synthetic CLDs | | | | | Reference CLD |
| --- | --- | --- | --- | --- | --- | --- |
| | Mean | Std. | Min. | Median | Max. | |
| Num. of Variables | 3.75 | 0.91 | 2 | 4 | 7 | 4 |
| Num. of Causal Links | 3.88 | 1.77 | 1 | 4 | 9 | 5 |
| Num. of Cycles | 1.14 | 1.11 | 0 | 1 | 5 | 2 |
| Density | 0.4 | 0.17 | 0.1 | 0.33 | 1 | 0.4 |
| Transitivity | 0.05 | 0.13 | 0 | 0 | 1 | 0.25 |
| Connectivity | 0.84 | 0.38 | 0 | 1 | 1 | 1 |

For text data diversity, we examine vocabulary statistics following [9]. Out of 7,508 variables, there were 473 unique variable names. S1 Figure illustrates the 15 most frequently occurring variable names within the synthetic CLDs, highlighting key concepts commonly represented in the causal graph generated by the LLM.

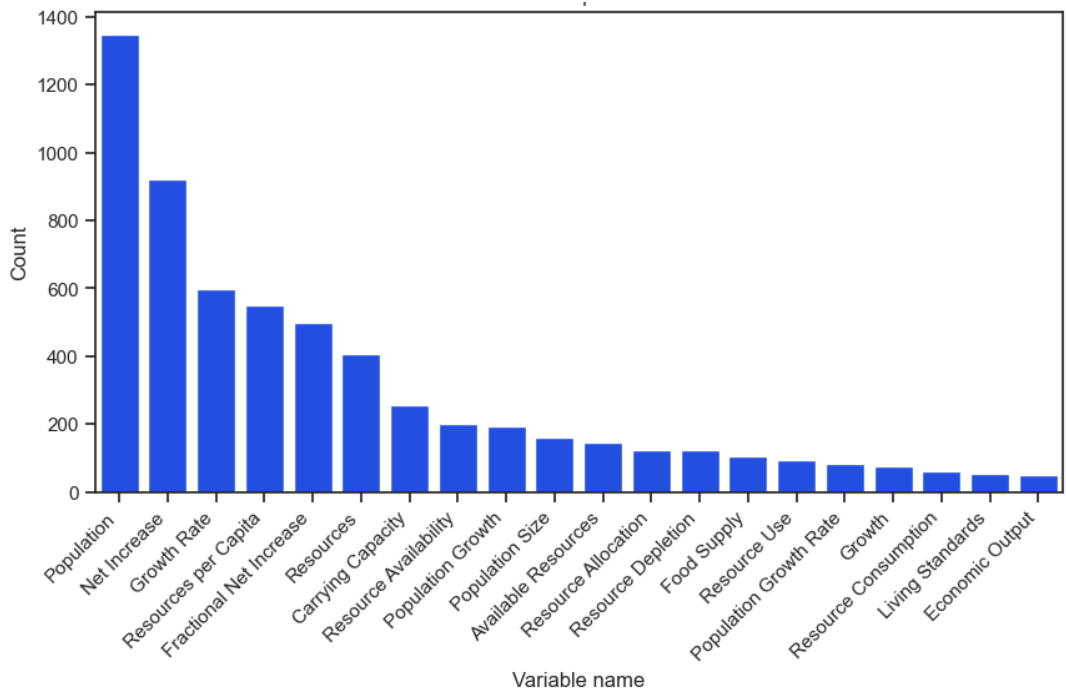

**S1 Figure.** The 15 most frequently occurring variable names in the 2,000 synthetic CLDs